# Unsupervised deep learning for individualized brain functional network identification

Hongming Li, Yong Fan

Center for Biomedical Image Computing and Analytics (CBICA), Department of Radiology, Perelman School of Medicine, University of Pennsylvania, Philadelphia, PA, 19104, USA

**Abstract.** A novel unsupervised deep learning method is developed to identify individual-specific large scale brain functional networks (FNs) from resting-state fMRI (rsfMRI) in an end-to-end learning fashion. Our method leverages deep Encoder-Decoder networks and conventional brain decomposition models to identify individual-specific FNs in an unsupervised learning framework and facilitate fast inference for new individuals with one forward pass of the deep network. Particularly, convolutional neural networks (CNNs) with an Encoder-Decoder architecture are adopted to identify individual-specific FNs from rsfMRI data by optimizing their data fitting and sparsity regularization terms that are commonly used in brain decomposition models. Moreover, a time-invariant representation learning module is designed to learn features invariant to temporal orders of time points of rsfMRI data. The proposed method has been validated based on a large rsfMRI dataset and experimental results have demonstrated that our method could obtain individual-specific FNs which are consistent with well-established FNs and are informative for predicting brain age, indicating that the individual-specific FNs identified truly captured the underlying variability of individualized functional neuroanatomy.

## 1 Introduction

Large scale brain functional networks (FNs) derived from resting-state functional magnetic resonance imaging (rsfMRI) data have been widely used for exploring functional neuroanatomy and functional connectome of the human brain. There is increasing evidence of marked individual heterogeneity in FNs and a series of studies have established that individual-specific FNs could effectively characterize cognition [1], brain development [2], and brain disorders [3], highlighting the need for tools that identify individual-specific FNs reliably and effectively.

A number of methods have been developed to identify individual-specific FNs by modeling them as latent factors of rsfMRI data in an unsupervised learning framework. Most of the existing methods adopt a two-step inference strategy that first estimates group FNs based on rsfMRI data from a group of individuals and then use the group FNs to infer individual-specific FNs under different assumptions, including group independent component analysis (ICA) [4], dual regression [5], group information guided ICA [6], multi-session hierarchical Bayesian model [1], and deep learning based methods built upon restricted Boltzmann machines [7] and deep belief networks (DBNs) [8]. However, these methods do not necessarily yield an optimal solution in that their individualized FNs are not optimized in the same way as their group level counterparts.

Several alternative methods have been developed to identify individual-specific FNs of different individuals jointly and enforce inter-individual correspondence of FNs using spatial group sparsity regularization [9] or regularizations based on certain assumptions about statistical distribution of loadings of corresponding FNs of different individuals [10, 11]. The DBNs have also been utilized to identify FNs of multiple individuals jointly [12]. However, all these methods cannot directly make inference for new individuals and are computationally expensive.

Convolutional neural networks (CNNs) have also been used to predict individual-specific default mode network (DMN) in a supervised learning setting [13], with DMNs obtained by conventional brain decomposition methods as ground truth. However, its performance may be limited by the silver standard ground truth adopted.

To overcome limitations of the existing methods, we develop a novel unsupervised learning framework, the first of its kind, to identify individual-specific FNs from rsfMRI data by leveraging conventional brain decomposition models and deep learning techniques. Particularly, CNNs with an Encoder-Decoder architecture are adopted to identify individual-specific FNs from rsfMRI data in an end-to-end learning fashion by optimizing their data fitting and sparsity regularization terms that are commonly used in brain decomposition models. Since time points of rsfMRI data of two different individuals do not align in temporal direction with each other, feeding 4D rsfMRI data into CNNs as multi-channel 3D data will yield instable features learning. Therefore, we develop a novel time-invariant representation learning module to make features learned by CNNs invariant to temporal orders of time points of rsfMRI data. Instead of enforcing the inter-individual comparability of FNs explicitly, the Encoder-Decoder CNNs facilitate the inter-individual comparability of FNs implicitly by the fact that voxels with similar functional profiles will be assigned to the same FN. A trained model of the Encoder-Decoder CNNs can be applied to rsfMRI data of a new individual to identify its FNs in one forward pass. We have evaluated the proposed method using rsfMRI data of 969 individuals from the Philadelphia Neurodevelopmental Cohort (PNC) [14] and experimental results have demonstrated that our method could obtain individual-specific FNs which coincide with well-established large scale FNs and are informative for predicting brain age, indicating that the individual-specific FNs identified by our method truly captured the variability of individualized functional neuroanatomy during brain development.

## 2    Methods

To characterize the individualized functional neuroanatomy under an unsupervised learning setting, we design an Encoder-Decoder architecture of CNNs, consisting of an Encoder-Decoder U-Net [15] module to identify FNs at individual level and a time-invariant representation learning module to learn features invariant to temporal orders of time points of rsfMRI data. Data fitting and spatial sparsity regularization terms are utilized to drive the CNNs to identify functionally coherent and spatially compact FNs. The overall framework is schematically illustrated in Fig. 1(a).

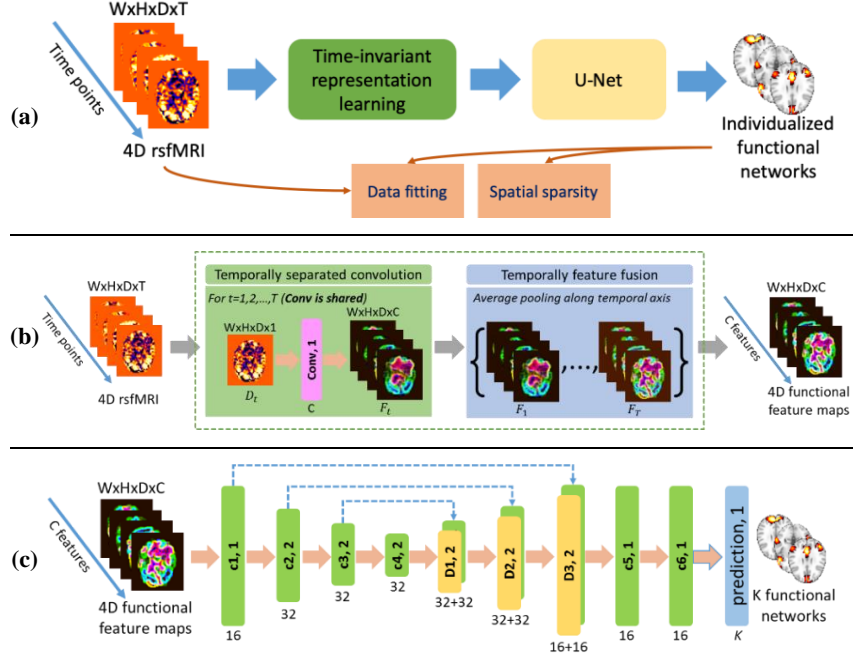

**Fig. 1.** Schematic diagram of the unsupervised deep learning model for identifying individual-specific functional networks: (a) overall network architecture, (b) time-invariant representation learning module, (c) U-Net module for the prediction of FNs.

### 2.1 Functional brain decomposition based on matrix factorization

Given rsfMRI data $X^i \in R^{T \times S}$ of individual $i$, consisting of $S$ voxels and $T$ time points, a matrix factorization based decomposition model tries to identify $K$ FNs $V^i = (V^i_{k,s}) \in R^{K \times S}$ and their corresponding time courses $U^i = (U^i_{t,k}) \in R^{T \times K}$, so that the original rsfMRI data can be approximated by two low-rank matrices $X^i \approx U^i V^i$. To favor FNs that do not contain anti-correlated functional units and are spatially compact and overlapped, non-negative constraint and spatial sparsity regularization are usually applied to $V^i$ [9], and the FNs can be obtained by optimizing

$$\min_{U^i, V^i} \|X^i - U^i V^i\|_F^2 + \lambda \sum_{j=1}^{K} \frac{\sum_{s=1}^{S} |V^i_{j,s}|}{\sqrt{\sum_{s=1}^{S} (V^i_{j,s})^2}}, \qquad s.t. V^i \in R_+, \qquad (1)$$

where the first term is a data fitting term, the second term is the Hoyer regularization for spatial sparsity [16], and $\lambda$ is the trade-off parameter.

An alternative optimization method is usually adopted to optimize Eq. (1) [17]. When $V^i$ is fixed, $U^i$ can be calculated analytically as $U^i = X^i (V^i)^T (V^i (V^i)^T)^{-1}$.

Substituting this expression for $U^i$ in Eq. (1), we have

$$\min_{V^i} \|X^i - X^i(V^i)^T(V^i(V^i)^T)^{-1}V^i\|_F^2 + \lambda \sum_{j=1}^{K} \frac{\sum_{s=1}^{S}|V_{j,s}^i|}{\sqrt{\sum_{s=1}^{S}(V_{j,s}^i)^2}}, \quad s.t. V^i \in R_+. \quad (2)$$

Instead of optimize $V^i$ using the multiplicative weight update method [17], we propose to train a deep network of CNNs to estimate $V^i$ directly from the input rsfMRI data as shown in Fig. 1(a).

### 2.2 Functional brain decomposition based on deep learning

Given a group of $n$ individuals, each having 4D rsfMRI data $I^i \in R^{W \times H \times D \times T}$, $i = 1, \ldots, n$, where $W$, $H$, and $D$ are width, height, and depth of each 3D volume respectively and $T$ is the number of time points, we train a deep network of 3D CNNs $M_{\theta_c}(I) = V_I$ with convolutional parameters $\theta_c$, which takes the 4D fMRI data $I$ as input and identifies its corresponding FNs $V_I$ as output. We use rsfMRI data from $n$ individuals as training data to minimize the loss function in Eq. (2) for learning the convolutional parameters $\theta_c$ (when calculate the loss function, $I^i$ is flatten across the spatial dimension, transposed and reshaped as a matrix with size $T \times S$, where $S = W \times H \times D$. The same transformation is applied to $V_I$ to get the reshaped matrix with size $K \times S$). As FNs are optimized as formulated by Eq. (2), the deep learning model is optimized in a self-supervised way. Once $\theta_c$ is optimized, the deep learning model can be used to predict individualized FNs for different individuals.

The deep network includes a representation learning module and one U-Net [15] module for identifying FNs, as illustrated in Fig. 1(a). The representation learning module (with details in next paragraph) extracts time-invariant feature maps from the input rsfMRI using 3D CNNs, and the feature maps are used as input to the U-Net module. As shown in Fig. 1(c), the U-Net module consists of one convolutional layer with 16 filters, three convolutional layers with 32 filters and a stride of 2, three deconvolutional layers with 32, 32, and 16 filters and a stride of 2, and finally 2 convolutional layers with 16 filters. LeakyReLu activation and batch normalization [18] are used for all the convolutional and deconvolutional layers. One output convolutional layer is used to predict individualized FNs and the number of output channels is $K$. Sigmoid activation is used for the output layer so that the output is non-negative. Each output channel is linearly scaled so that its maximum equals to 1 before calculating the loss function. The kernel size in all layers are set to 3×3×3.

One major challenge to the deep learning model is to learn features invariant to temporal orders of time points of rsfMRI data that do not align in temporal direction across different scans, no matter they are from different individuals or different sessions of the same individual. As the rsfMRI data from different scans cannot be compared directly, it is not feasible to feed the rsfMRI data into 3D CNNs straightforwardly as a multi-channel input. Inspired by the study [19] which has demonstrated that replication of FNs can be obtained through averaging of selected fMRI time frames, we propose a time-invariant representation learning module to learn features invariant to temporal orders of time points of rsfMRI scans and facilitate the identification of FNs by the following U-Net module. As shown in Fig. 1(b), the representation learning module consists of two parts referred as temporally separated convolu-tion and temporal feature fusion. The temporally separated convolution part is a 3D convolutional layer with $C$ filters and a stride of 1, which is applied to each time point (a 3D volume with size of $W \times H \times D$) of the rsfMRI scan and output $C$ feature maps (with size of $W \times H \times D \times C$) for each time point. The temporal feature fusion part is an element-wise average pooling layer which outputs $C$ average feature maps of all the time points. In the present study, $C$ is set to 16 and the kernel size of the convolutional layer is set to 3×3×3, and LeakyReLu activation and batch normalization are also adopted. While the convolution learning part is expected to capture the co-activation patterns of different brain regions by different feature channels at each time point, the feature fusion part summarizes the patterns obtained from all the time points. This module is plugged into the model just before the U-Net module, and the whole model can be optimized in an end-to-end fashion.

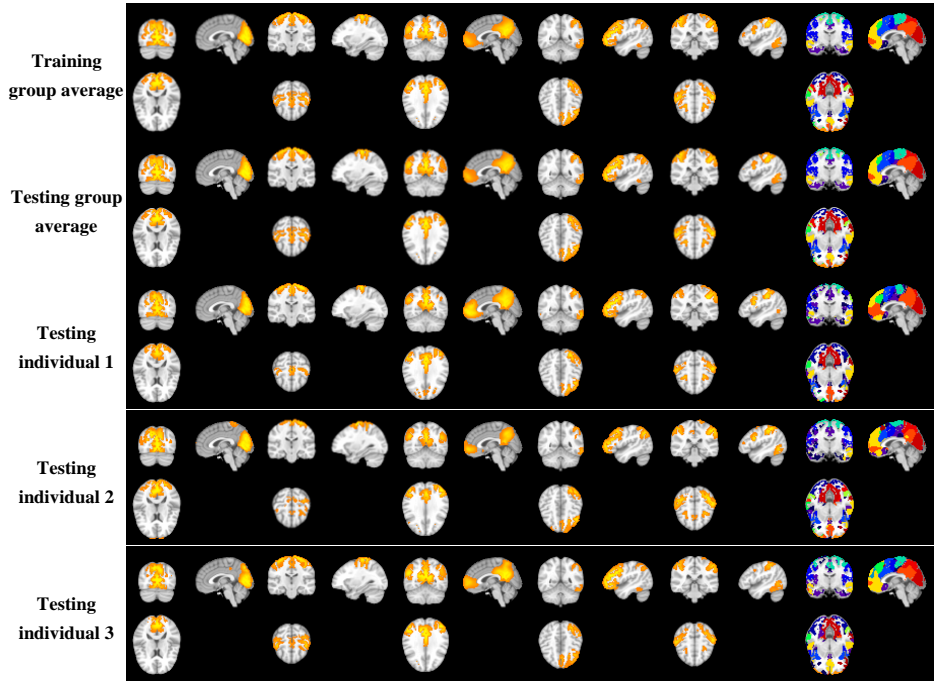

**Fig. 2.** Functional networks identified by the proposed method. Group average FNs of training individuals, testing individuals, and individualized FNs of three randomly selected testing individuals are demonstrated from row 1 to 5. Columns 1 to 5 refer to different FNs, and column 6 shows FN maps illustrating all 17 FNs, in which each color indicates one specific FN and the same color indicates corresponding FNs across individuals.

## 3   Experimental results

We evaluated our method based on rsfMRI data of 969 individuals from the PNC dataset [14] (ages from 8 to 22). The fMRI data were preprocessed using an optimized procedure, including slice timing, confound regression, and band-pass filtering [20].

Each individual's rsfMRI scan has 118 time points after preprocessing. We randomly selected 500 individuals as training data to optimize the proposed deep learning model, and the remaining 469 individuals were used as testing data. It is worth noting that no ground truth FNs are required for training the proposed model.

In the present study, the number of FNs was set to 17 in order to facilitate a direct comparison with the well-established FNs [21]. Our model was implemented using Tensorflow. Adam optimizer was adopted to optimize the network, the learning rate was set to $1 \times 10^{-4}$, the batch size was set to 1, and the number of iterations was set to 30000 during training. One NVIDIA TITAN Xp GPU was used for training and testing. We set $\lambda = 0.001$ empirically in our experiments.

### 3.1 Individualized brain functional networks

The FNs identified by the proposed method are illustrated in Fig. 2. Group average FNs of training individuals and testing individuals, as well as individualized FNs of three randomly selected individuals are illustrated in rows of 1 to 5 respectively. Our method successfully identified both spatially localized and distributed FNs that support primary and high order brain functions, such as visual network, somatomotor network, default mode network, fronto-parietal network, dorsal attention network as shown from columns of 1 to 5 respectively. FN maps illustrating all 17 FNs are shown in column 6, where each color corresponds to one FN and each voxel is assigned to the FN whose coefficient had the largest value across all FNs. The group average FNs and FN map from training and testing individuals are visually similar as demonstrated in the 1st and 2nd row, indicating that the proposed model generalize well from the training data to the testing data and highlighting generalization ability of the proposed unsupervised learning method. Furthermore, inter-individual differences can be clearly observed in the individualized FNs and FN maps shown in rows of 3 to 5 though the shapes of FNs are consistent across individuals, indicating the proposed method captured the inherent differences of individualized functional neuroanatomy. On average, it took $4.75 \pm 0.183$ seconds to use the trained model to identify 17 FNs for each testing individual, highlighting its computational efficiency.

### 3.2 Brain age prediction based on individualized functional networks

As no ground truth is available for FNs derived from rsfMRI data, we evaluated the individualized FNs for predicting brain age based on their FN loadings that characterize functional network topography (i.e., shape and location) with an assumption that individualized FNs could characterize the changes of functional neuroanatomy associated with brain development.

Particularly, loadings of the individualized FNs (17 FNs) of each testing individual obtained by the proposed method were flattened and concatenated as one vector and used as features to predict its brain age using ridge regression. 469 testing individuals were used for the brain age prediction and the prediction performance was evaluated using a two-fold cross-validation. A nested two-fold cross-validation was adopted to identify the optimal hyper-parameter in the ridge regression model. The whole prediction procedure was repeated 100 times, and the prediction accuracy was evaluated as the Pearson correlation coefficient and mean absolute error (MAE) between the predicted brain age and the chronological age. We compared the prediction performance of brain age prediction models built on FNs obtained by the proposed method and those obtained by a nonnegative matrix factorization (NMF) based brain decomposition model (https://github.com/hmlicas/Collaborative_Brain_Decomposition) [9].

Age distribution of all 469 individuals and the brain age prediction performance of the individualized FNs are illustrated in Fig. 3. The prediction models based on FNs obtained by the proposed method predicted the brain age with higher accuracy (correlation coefficient: $0.479 \pm 0.028$ and MAE: $2.23 \pm 0.051$ years), significantly better than that based on FNs obtained by the NMF based decomposition model (denoted as FNs_NMF in Fig. 3, correlation coefficient: $0.367 \pm 0.038$ and MAE: $2.39 \pm 0.045$ years). The performance improvement indicated that our method could better characterize inter-individual variability as no specific spatial regularization was required to maintain inter-individual correspondence. To make sure that the prediction was not driven by the confounding influences of in-scanner head motion of rsfMRI data or sex, we also carried out the same prediction procedure using motion measures obtained from the rsfMRI preprocessing procedure and sex as features. The prediction models of motion and sex had significantly worse performance than those built upon individualized FNs, with a correlation coefficient of $0.129 \pm 0.040$ and a MAE of $2.56 \pm 0.022$ years. These results indicated that individualized FNs obtained by our method could better capture the individualized functional neuroanatomy underlying the brain development.

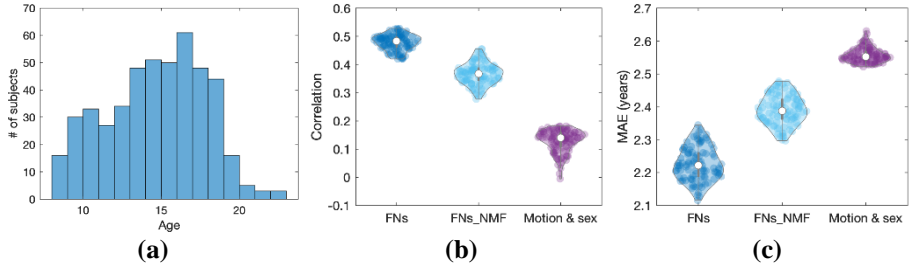

**Fig. 3.** Brain age prediction based on the FNs identified by the proposed method. (a) Age distribution of 469 testing individuals, (b) correlation and (c) MAE measures between predicted brain age and chronological age from 100 repetitions of two-fold cross-validation by different models.

## 4 Conclusions

In this study, we have developed an unsupervised learning model to identify individual-specific functional networks with inter-individual correspondence. Our method is built upon an Encoder-Decoder network of CNNs, and its optimization is driven in a self-supervision fashion by a loss consisting of data fitting and spatial sparsity regularization terms that have been successfully used in a matrix factorization based brain decomposition model [9]. Moreover, a novel time-invariant representation learning module is proposed to learn features invariant to temporal orders of time points of rsfMRI scans. Experimental results based on a large rsfMRI dataset have demonstrated that our method could efficiently identify individualized FNs to capture the differences of functional neuroanatomy across individuals during brain development.

**References**

1. Kong, R., et al., *Spatial Topography of Individual-Specific Cortical Networks Predicts Human Cognition, Personality, and Emotion.* Cereb Cortex, 2019. **29**(6): p. 2533-2551.
2. Cui, Z., et al., *Individual Variation in Functional Topography of Association Networks in Youth.* Neuron, 2020.
3. Wang, D., et al., *Individual-specific functional connectivity markers track dimensional and categorical features of psychotic illness.* Mol Psychiatry, 2018.
4. Calhoun, V.D., et al., *A method for making group inferences from functional MRI data using independent component analysis.* Hum Brain Mapp, 2001. **14**(3): p. 140-51.
5. Nickerson, L.D., et al., *Using Dual Regression to Investigate Network Shape and Amplitude in Functional Connectivity Analyses.* Frontiers in Neuroscience, 2017. **11**.
6. Du, Y. and Y. Fan, *Group information guided ICA for fMRI data analysis.* Neuroimage, 2013. **69**: p. 157-97.
7. Hjelm, R.D., et al., *Restricted Boltzmann machines for neuroimaging: an application in identifying intrinsic networks.* Neuroimage, 2014. **96**: p. 245-60.
8. Dong, Q., et al., *Modeling Hierarchical Brain Networks via Volumetric Sparse Deep Belief Network (VS-DBN).* IEEE Trans Biomed Eng, 2019.
9. Li, H., T.D. Satterthwaite, and Y. Fan, *Large-scale sparse functional networks from resting state fMRI.* Neuroimage, 2017. **156**: p. 1-13.
10. Abraham, A., et al., *Extracting brain regions from rest fMRI with total-variation constrained dictionary learning.* Med Image Comput Comput Assist Interv, 2013. **16**(Pt 2): p. 607-15.
11. Harrison, S.J., et al., *Large-scale probabilistic functional modes from resting state fMRI.* Neuroimage, 2015. **109**: p. 217-31.
12. Zhang, S., et al., *Discovering hierarchical common brain networks via multimodal deep belief network.* Med Image Anal, 2019. **54**: p. 238-252.
13. Zhao, Y., et al., *4D Modeling of fMRI Data via Spatio-Temporal Convolutional Neural Networks (ST-CNN).* IEEE Transactions on Cognitive and Developmental Systems, 2019: p. 1-1.
14. Satterthwaite, T.D., et al., *Neuroimaging of the Philadelphia neurodevelopmental cohort.* Neuroimage, 2014. **86**: p. 544-53.
15. Ronneberger, O., P. Fischer, and T. Brox. *U-net: Convolutional networks for biomedical image segmentation.* in *International Conference on Medical image computing and computer-assisted intervention.* 2015. Springer.
16. Hoyer, P.O., *Non-negative matrix factorization with sparseness constraints.* Journal of machine learning research, 2004. **5**(Nov): p. 1457-1469.
17. Ding, C.H., T. Li, and M.I. Jordan, *Convex and semi-nonnegative matrix factorizations.* IEEE transactions on pattern analysis and machine intelligence, 2008. **32**(1): p. 45-55.
18. Ioffe, S. and C. Szegedy, *Batch normalization: Accelerating deep network training by reducing internal covariate shift.* arXiv preprint arXiv:1502.03167, 2015.

19. Liu, X. and J.H. Duyn, *Time-varying functional network information extracted from brief instances of spontaneous brain activity.* Proc Natl Acad Sci U S A, 2013. **110**(11): p. 4392-7.
20. Ciric, R., et al., *Benchmarking of participant-level confound regression strategies for the control of motion artifact in studies of functional connectivity.* Neuroimage, 2017. **154**: p. 174-187.
21. Yeo, B.T., et al., *The organization of the human cerebral cortex estimated by intrinsic functional connectivity.* J Neurophysiol, 2011. **106**(3): p. 1125-65.